%% file: emnlp2020.tex
\documentclass[11pt,a4paper]{article}
\usepackage[hyperref]{emnlp2020}
\usepackage{times}
\usepackage{latexsym}
\usepackage{tabularx}
\usepackage{booktabs}

\newcommand{\eat}[1]{}
\newcommand{\red}[1]{\textcolor{red}{#1}}
\newcommand{\blue}[1]{\textcolor{blue}{#1}}

\newcommand{\teal}[1]{\textcolor{teal}{#1}}

\usepackage{quoting}
\newenvironment{myquote}{                   
  \parskip 0mm \begin{quoting}[vskip=0mm,leftmargin=2mm]}{
\end{quoting}}
\newenvironment{des}{                 
     \parskip 0cm \begin{list}{}{\parsep 0cm \itemsep 0cm \topsep 0cm}}{
       \end{list}} 

\usepackage{microtype}

\aclfinalcopy 
\def\aclpaperid{2281} 

\usepackage{graphicx}
\usepackage{subcaption}
\usepackage{amsmath}
\usepackage{mathtools}
\usepackage{amsfonts}

\newcommand{\cut}[1]{}

\newcommand{\eqasc}{\texttt{eQASC}}
\newcommand{\eqascpert}{\texttt{eQASC-perturbed}}
\newcommand{\eobqa}{\texttt{eOBQA}}

\title{
Learning to Explain: Datasets and Models for Identifying Valid Reasoning Chains in Multihop Question-Answering}

\author{Harsh Jhamtani $^1$ \qquad Peter Clark $^2$ \\
$^1$ School of Computer Science, Carnegie Mellon University\\
$^2$ Allen Institute for AI \\
\tt{jharsh@cs.cmu.edu, peterc@allenai.org}
}

\date{}

\begin{document}

\maketitle

\begin{abstract}
 Despite the rapid progress in multihop question-answering (QA), models still have trouble explaining
 {\it why} an answer is correct, with limited explanation training data available to learn from.
 To address this, we introduce three explanation datasets in which explanations formed from corpus facts are annotated. Our first dataset, \eqasc{}, contains over 98K  explanation annotations for the multihop question answering dataset QASC,
 and is the first that annotates {\it multiple} candidate explanations for each answer. 
 The second dataset \eqascpert{} is constructed by crowd-sourcing perturbations (while preserving their validity) of a subset of explanations in QASC, to test consistency and generalization of explanation prediction models. The third dataset \eobqa{} is constructed by adding explanation annotations to the OBQA dataset 
 to test generalization of models trained on \eqasc{}. 
 We show that this data can be used to significantly improve explanation quality (+14\% absolute F1 over a strong retrieval baseline) using a BERT-based classifier, but still behind the upper bound, offering a new challenge
 for future research.
 We also explore a delexicalized chain representation in which repeated noun phrases are replaced by variables, thus turning them into {\it generalized reasoning chains} (for example: "X is a Y" AND "Y has Z" IMPLIES "X has Z"). We find that generalized chains maintain performance while also being more robust to certain perturbations.\footnote{Code and datasets can be found  at \url{ https://allenai.org/data/eqasc}}
 \end{abstract}

\setlength{\belowdisplayskip}{4pt} \setlength{\belowdisplayshortskip}{4pt}
\setlength{\abovedisplayskip}{4pt} \setlength{\abovedisplayshortskip}{4pt}

\input{texfiles/intro.tex}

\input{texfiles/related.tex}
\input{texfiles/data.tex}

\input{texfiles/method.tex}

\input{texfiles/experiments.tex}

\input{texfiles/analysis.tex}

\input{texfiles/conclusions.tex}

\paragraph{Acknowledgements}
We thank Ashish Sabharwal, Tushar Khot, Dirk Groeneveld, Taylor Berg-Kirkpatrick, and anonymous reviewers for useful comments and feedback. We thank Michal Guerquin for helping with the QA2D tool. This work was partly carried out when HJ was interning at AI2. HJ is funded in part by a Adobe Research Fellowship.

\bibliography{emnlp2020}
\bibliographystyle{acl_natbib}

\input{appendix.tex}

\end{document}

%% file: texfiles/intro.tex
\section{Introduction}

\begin{figure}[t]
{
\centerline{
 \fbox{%
  \parbox{1\columnwidth}{\small
    {\bf Q:} What can cause a forest fire? \\
    \hspace*{4mm} (1) rain (2) static electricity (3) microbes (4) ... \\
            {\bf A:} static electricity \\
            {\bf Q+A} (declarative): Static electricity can cause a forest fire. \\
 \ \\
     \vspace{1mm}  
\underline{\bf Explanation (reasoning chain):} [positive (valid)] \\
\hspace*{7mm}  {\it \red{Static electricity} can cause \blue{sparks}} \hspace*{3mm} // (from corpus) \\
{\bf AND} {\it \blue{Sparks} can start a \teal{forest fire}} \hspace*{10.5mm} // (from corpus) \\
\hspace*{3mm} {\bf $\rightarrow$} {\it \red{Static electricity} can cause a \teal{forest fire}}  \hspace{5mm} // (Q+A) \\
    \ \\
    \vspace{1mm}
    \underline{\bf Explanation (Generalized reasoning chain, GRC):} \\
{\it {\bf X} can cause {\bf Y} {\bf AND} {\bf Y} can start {\bf Z} {\bf $\rightarrow$} {\bf X} can cause {\bf Z}}
}}}
\caption{
Our datasets contain annotated (valid and invalid) {\it reasoning chains} in support of an answer, allowing explanation classifier models to be trained and applied. We also find that using a variabilized version of the chains improves the models' robustness.} 
  \label{example}}
\end{figure}

\begin{figure*}[t]
    \centering
    \includegraphics[width=0.85\textwidth]{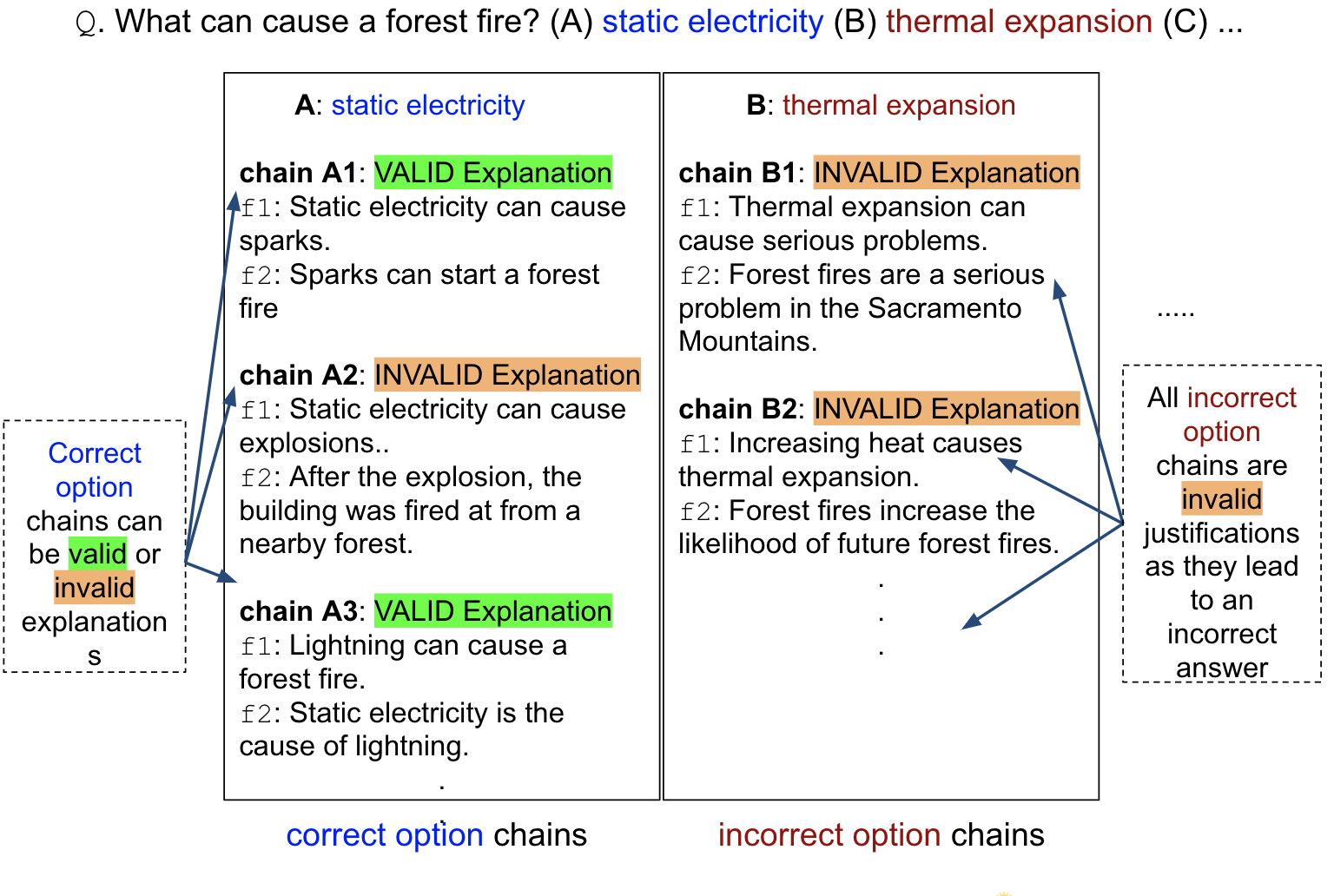}
    \vspace{-5mm}
    \caption{
    QASC contains multiple-choice questions, plus one gold (valid) reasoning chain for the correct answer. To find valid reasoning chains, we first generate candidates for each answer option using a 2-step retrieval process (Section~\ref{retrieval}). 
    We then collect annotations for the correct answer option chains to train and evaluate models to detect valid reasoning chains.
     (Above, chains A1 and A3 are valid, while A2, B1, and B2 are invalid).
}
\label{fig:setup}
\end{figure*}

While neural systems have become remarkably adept at question answering (QA), e.g., \cite{Clark2017SimpleAE},
their ability to {\it explain} those answers remains limited. This creates
a barrier for deploying QA systems in practical settings, and limits
their utility for other tasks such as education and tutoring, where
explanation plays a key role.
This need has become particularly important with 
{\it multihop question-answering}, where multiple facts are
needed to derive an answer\cut{\cite{OpenBookQA2018,welbl2018constructing}}. In this context, seeing a
{\it chain of reasoning} leading to an answer,  can help a user assess an answer's validity.
Our research here contributes to this goal.

We are interested in questions where the decomposition into
subquestions - hence the explanation structure - is {\it not evident
from the question}, but has to be found.
For example, ``Does a suit of armor conduct electricity?'' might be
answered (hence explained) by first identifying what material armor is made of,
even though the question itself does not mention materials. This contrasts with earlier multihop QA datasets, e.g., HotpotQA \cite{yang2018hotpotqa}, where the explanation structure is evident in the question itself. For example, ``What nationality was James Miller's wife?'' implies a chain of reasoning to first finds Miller's wife, then her nationality. Such cases are easier but less representative of natural questions. Multihop datasets of the kind where explanation structure is not evident  include OpenBookQA \cite{OpenBookQA2018} and more
recently QASC \cite{khot2019qasc}. However, although providing QA pairs,
these datasets provide limited explanation information.
OpenBookQA does not come with any explanation data, and QASC only
provides a single gold explanation for each answer, 
while in practice there may be multiple valid explanations.

To alleviate this lack of data, we contribute three new datasets:
The first (and largest) is \eqasc{}, containing annotations on over 98K 
candidate explanations for the QASC dataset, including on {\it multiple} (typically 10)
possible explanations for each answer, including both valid and invalid explanations. The second, \eqascpert{}, contains
semantically invariant perturbations of a subset of QASC explanations,
for better measuring the generality of explanation prediction models.
Finally \eobqa{} adds adding explanation annotations to the OBQA test set,
to further test generality of models trained on \eqasc{}.
In addition, we use these datasets to build models for detecting
valid explanations, to establish baseline scores. Finally, we
explore a delexicalized chain representation in which repeated
noun phrases are replaced by variables, thus turning them
into {\it generalized reasoning chains}, as illustrated in Figure~\ref{example}.
We find that generalized chains maintain performance while also being more
robust to perturbations, suggesting a promising avenue for further research.

%% file: texfiles/related.tex
\section{Related Work}

In the context of QA, there are multiple notions of explanation/justification,
including showing an authoritative, answer-bearing sentence \cite{perez2019finding},
a collection of text snippets supporting an answer \cite{DeYoung2020ERASERAB},
an attention map over a passage \cite{Seo2016BidirectionalAF}, a synthesized phrase
connecting question and answer \cite{Rajani2019ExplainYL}, or the syntactic pattern
used to locate the answer \cite{Ye2020TeachingMC,Hancock2018TrainingCW}.
These methods are primarily designed for answers to ``lookup'' questions, to
explain where and how an answer was found in a corpus.

For questions requiring inference, the focus of this paper,
an explanation is often taken as the chain of steps (typically sentences)
leading to an answer. HotpotQA's support task goes partway towards this
by asking for answer-supporting sentences (but not how they combine) \cite{yang2018hotpotqa}.
The R4C dataset takes this further, annotating how (and which) HotpotQA supporting
sentences chain together \cite{Inoue2019RCQEDEN}. However, in HotpotQA and R4C, the
decomposition (hence structure of the explanation) is evident in the
question \cite{OpenBookQA2018}, simplifying the task. More recently, multihop
datasets where the decomposition is not evident have appeared, e.g.,
WikiHop \cite{welbl2018constructing}, OBQA \cite{OpenBookQA2018}, and
QASC \cite{khot2019qasc}, posing more a realistic explanation challenge.
However, explanation annotations are sparse (only QASC contains a
single gold explanation per question), limiting their support for
both training and evaluation of explanation systems, hence motivating this work.

Finally there are a few human-authored explanation datasets.
e-SNLI \cite{camburu2018snli} adds crowdsourced explanations to SNLI entailment problems
\cite{Bowman2015ALA}, and CoS-E \cite{Rajani2019ExplainYL} adds explanations for 
CommonsenseQA questions \cite{Talmor2019CommonsenseQAAQ}. This work
differs from ours in two ways. First, the authored explanations are
single-hop, directly linking a question to an answer. Second, the
datasets were primarily designed for (explanation) language generation,
while our goal is to assemble explanations from an authoritative
corpus so that they have credible provenance.

Our work is quite different from prior work focusing on textual entailment. Our goal is not to decide if a sentence is entailed, but to identify a valid explanation for why. For example, a SciTail \cite{khot2018scitail} model may predict that ``metals conduct heat" entails ``a spoon transmits energy", but not offer an explanation as to why. Our work fills this gap by providing an explanation (e.g., ``spoon is made of metal", ``heat is energy" from a larger retrieved context). Similarly, another entailment-based dataset is FEVER \cite{thorne2018fact}, testing where a larger context entails a claim. However, the FEVER task requires finding a context sentence that simply paraphrases the claim, rather than a reasoned-based explanation from more general statements - the aim of this work. 

%% file: texfiles/data.tex
\section{Explanation Datasets \label{datasets}}

We now present our new datasets, first describing how we construct candidate chains for each QA pair, and then how they were annotated.

\begin{table}[]
{\small
    \centering
    \begin{tabular}{@{}rccc@{}}
        \multicolumn{1}{c}{} & \bf Train  & \bf Dev  & \bf Test  \\ \toprule
        Total number of questions & 8134 & 926 & 920 \\ 
        Total no. of chains tagged & 80449 & 9190 & 9141  \\ 
        No. of valid chains & 21551 & 2186 & 2210 \\ 
        No. of invalid chains & 58898 & 7004 & 6931 \\ 
        \bottomrule
    \end{tabular}
    \caption{Summary statistics for \eqasc{}, the annotated chains for the correct answers in QASC. Each chain is tagged by three annotators, and we use majority judgement. 
    } 
    \label{tab:annotation}
    }
\end{table}

\subsection{Task Definition}

We consider the task where the input is a question Q, (correct) answer A, and a corpus of sentences $T$. The (desired) output is a valid reasoning chain, constructed from sentences in $T$,
that supports the answer. 
We define a reasoning chain as a sequence of sentences $C = [s_1,...,s_n]$ plus a conclusion sentence $H$, and a {\it valid} reasoning chain as one where $C$ {\it entails} $H$.
Following the textual entailment literature \cite{Dagan2013RecognizingTE}, we define entailment using human judgements rather than formally, i.e., $C$ entails $H$ if a person would reasonably
conclude $H$ given $C$. This definition directly aligns with our end-goal,
namely to provide users with a credible reason that an answer is correct.

For generating candidate chains $C$, we construct each $C$ from sentences in
the corpus $T$, as described below. Following the design of the QASC dataset,
we consider just 2-sentence chains, as this was the maximum chain
length used in its creation, although our approach could be extended to N-sentence chains.

\subsection{Candidate Chain Construction \label{retrieval}}

Given Q + A, we use the procedure described in \cite{khot2019qasc} to assemble candidate chains from $T$ (below). This procedure aims
to find plausible chains by encouraging word overlap:
\begin{des}
\item[(1)] Using
ElasticSearch (a standard retrieval engine),
retrieve K (=20 for efficiency) facts F1 from $T$ using
Q+A as the search query.
\item[(2)] For each 
fact f1 $\in$ F1, retrieve L (=4 to promote diversity) facts F2,
each of which contains at least one word from Q+A \textbackslash~f1 and
from f1 \textbackslash~Q+A;
\item[(3)] Remove [f1,f2] pairs that do not contain
any word from Q or A;
\item[(4)] Select the top M (here, =10) [f1,f2] pairs
sorted by the sum of their individual IR (ElasticSearch) scores.
\end{des}
Step (3) ensures that the chain contains at least some
mention of part of Q and part of A, a minimal requirement.
Step (4) imposes a preference for chains with greater
overlap with Q+A, and between f1 and f2. Note that
this procedure does not guarantee {\it valid} chains,
rather it only finds candidates that may be plausible
because of their overlap.
Some example chains are produced by this method are shown in Figure~\ref{fig:setup}.
In all our experiments, we use the QASC corpus\footnote{https://allenai.org/data/qasc}  as the corpus $T$, 
namely the same corpus of 17M cleaned up facts as used in \cite{khot2019qasc}.

\subsection{eQASC - Explanations for QASC \label{eqasc-section}}

The original QASC dataset includes only a single gold (valid) reasoning chain for
each correct answer, and no examples of invalid chains. To develop a richer
explanation dataset, suitable for both training and evaluation, we
generate eQASC as follows. First, we use the above algorithm 
to generate (up to) 10 candidate chains for each Q + correct answer option A pair.
This resulted in a total of {\bf 98780 chains} for QASC's 9980 questions.

We then use (Amazon Turk) crowdworkers to annotate each chain.
Workers were shown the question, correct answer, and reasoning chain, e.g.:
\vspace{2mm}

\centerline{
 \fbox{%
   \parbox{0.98\columnwidth}{\small
 Question: {\bf What is formed by rivers flowing over rocks?} \\
 Answer: {\bf soil} \\
 Because: \\
 \hspace*{5mm} fact 1: {\bf Rivers erode the rocks they flow over}, {\it and} \\
 \hspace*{5mm} fact 2: {\bf soil is formed by rocks eroding} 
}}}
\vspace{2mm}

\noindent
They were then asked if fact 1 and fact 2 {\it together} were a reasonable chain of reasoning for the answer,
and to promote thought were offered several categories of ``no'' answer: fact 1 alone, or fact 2 alone, or either alone, justified the answer; or the answer was not justified; or the question/answer did not make sense. (Two ``unsure'' categories were also offered but rarely selected). The full instructions to the workers are provided in the Appendix.
Each chain was annotated by 3 workers. To ensure quality, only AMT Masters level workers were used, and several
checks were performed:
First, for cases where at least two workers agreed, if a worker's annotations disagreed with the majority
unreasonably often (from inspection, judged as more than 25\% of the time),
then the worker was (paid but then) blocked, and his/her annotations redone.
Second, if a worker's distribution of labels among the six categories substantially deviated from other
workers (e.g., almost always selecting the same category), or if
his/her task completion time was unrealistically low, then his/her work was sampled and checked. If it
was of low quality then he/she again was (paid and) blocked, and his/her annotations redone.
Pairwise agreement was 74\% (2 class) or 45\% (for all six subclasses), with
a Fleiss $\kappa$ (inter-annotator agreement) of 0.37 (``fair agreement'' \cite{fleiss}).
There was a majority agreement (using all six subclasses) of 84\%, again suggesting fair annotation quality.
For the final dataset, we adopt a conservative approach and treat the no majority agreement cases as invalid chains.
Summary statistics are in Table \ref{tab:annotation}.

\subsection{eQASC-perturbed  - Testing Robustness}
For a test of robustness of model for reasoning chain explanation detection, we also created \eqascpert{}, a dataset of valid eQASC reasoning chains, perturbed in a way so as to preserve their validity. To do this, we first randomly selected a subset of the valid reasoning chains from the test split of \eqascpert{}. We then asked crowdworkers to modify the chains by replacing a word or phrase shared between at least two sentences with a different word or phrase, and to make sure that the resulting new chain
remained valid. 
(e.g., {\it "amphibians" } became {\it "frogs"}, or {\it "eats other animals"} became 
{\it "consumes its prey"}).
\cut{
For example, given the chain:
\begin{myquote}
{\it Amphibians are important predators}  \\
AND {\it An animal that kills and eats other animals is called a predator} \\
$\rightarrow$ {\it Amphibians sometimes eat other animals} 
\end{myquote}
a crowdworker may rewrite it (changes underlined): 
\begin{myquote}
  {\it Amphibians are important predators} \\
AND {\it An animal that kills and \underline{consumes its prey} is called a predator}  \\
$\rightarrow$ {\it Amphibians sometimes \underline{consume their prey}} 
\end{myquote}
}
We collected {\bf 855 perturbed}, (still) valid reasoning chains in this way.

\subsection{eOBQA - Testing Generalization}

Finally, to further measure the generality of our model (without re-fine-tuning), we
created a smaller set of annotations for a different dataset, namely OBQA
(4-way multiple choice) \cite{OpenBookQA2018}. The original dataset has no explanations
and no associated corpus. Thus to generate explanations, we use
sentences from the QASC corpus, and annotate the top two (for all test questions) formed by the retrieval step (Section~\ref{retrieval}).
Note that for some questions, there may be no valid justification which can be formed from the corpus.
We followed the same annotation protocol as for \eqasc{} to have
crowd workers annotate the chains (Section~\ref{eqasc-section}). The resulting dataset
containing {\bf 998 annotated chains}, of which 9.5\% were marked as
valid reasoning explanations.

%% file: texfiles/method.tex
\section{Learning to Score Chains}

\begin{figure*}[]
    \centering
    \includegraphics[width=0.85\textwidth]{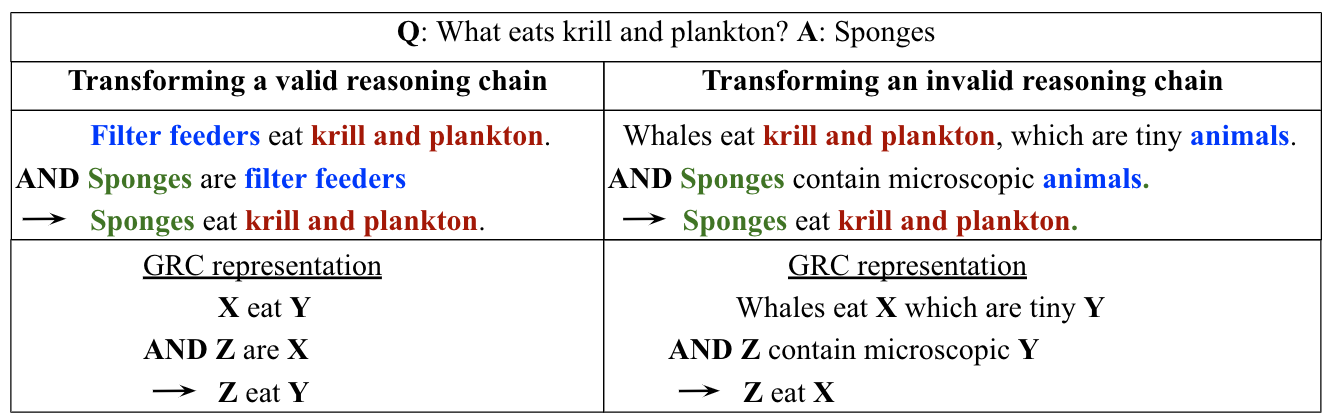}
    \caption{Generalized reasoning chains (GRCs) are formed by replacing repeated noun phrases
    with variables.} 
    \label{fig:variabilizing}
\end{figure*}

Our full approach to explaining an answer has two steps, namely candidate chain retrieval followed by chain scoring, to find the highest-ranked chain supporting an answer. For chain retrieval, we assume the same procedure described earlier to identify candidate chains. 
For chain scoring, we train a BERT-based model to distinguish valid chains from invalid ones, using the training data collected in the \eqasc{} dataset, as we now describe.

We evaluate using all three collected datasets.
We also evaluate two different ways of presenting the chain to the model to score
(both train and test): (a) in its original form (with Q+A flipped to a declarative sentence),
(b) in a generalized form, where 
repeated noun phrases are variabilized.
Our interest is how well a model can perform, both to assess practical
use and as a baseline for further improvement; and how the two different
chain representations impact performance.

\subsection{Chain Representation}

\noindent \textbf{Declarative form}
For a chain to support the answer to a question, we construct $H$ as a declarative form of the question + answer using standard QA2D tools, e.g., \cite{Demszky2018TransformingQA}. For example, for the question + answer ``What can cause a forest fire? Static electricity'', the hypothesis $H$  to be entailed by $C$ is ``Static electricity can cause a forest fire.''. An alternate representation for $H$ is to simply append answer to the end of the question. We did not observe any significant change in the best dev split performances on switching to the alternate representation described above. \\

\noindent \textbf{Generalized Reasoning Chains (GRC) \label{variabilization}:}
We observe that specific reasoning chains are often instantiations of more general patterns. For example, in Figure~\ref{example}, the specific explanation can be seen as an instantiation of the more general pattern ``X can cause Y'' AND ``Y can start Z'' IMPLIES ``X can cause Z''. We refer to such patterns as {\it Generalized Reasoning Chains} (GRCs). To encourage our model to recognize valid and invalid chains at the pattern level, we explore the following strategy: First, we transform candidate chains into generalized chains (GRCs) by replacing repeated noun phrases with variables (special tokens), a process known as delexicalization \cite{suntwal2019importance}. We then train and test the model using the GRC representation. We hypothesize that distinguishing a valid justification chain from an invalid one should not need typing information in most cases.

To identify the phrases to variabilize, (1) we first perform part-of-speech tagging on the sentences, and (2) extract candidate entities by identifying repeating nouns i.e. those which occur in at least two of the sentences in the chain (We stem the words before matching, and include any matching preceding determiners and adjectives into detected entities). e.g. `the blue whale is a mammal' and `the blue whale breathes..' leads to detection of 'the blue whale'). (3) Then, we assign a special token to each of the candidates, using a predefined set of unused tokens, which can be viewed as a set of variables. Some examples of GRCs are shown in Figure \ref{fig:variabilizing} and later in Figure~\ref{GRC-examples}, using {\tt X,Y,Z} as the special token set (As our models are BERT-based, we use {\tt unusedi} $i \in \{1,2,...\}$ to denote these tokens).

\subsection{Model Training \label{training}}

To distinguish valid from invalid chains, we fine-tune a pre-trained BERT model \cite{devlin2019bert} for scoring the possible explanation chains. We encode a chain $f1$ AND $f2$ $\rightarrow$ $H$ as: 
\begin{equation*}
\text{[CLS]} \enskip f1 \enskip \text{[SEP]} \enskip  f2 \enskip  \text{[SEP]} \enskip H    
\end{equation*} where [SEP] is a sentence boundary marker. Thereafter, we pass the chain through the BERT model (BERT-base-uncased). 
We employ a two layer feed-forward neural network with ReLU non-linearity, as a binary classifier  on the pooled [CLS] representation to predict valid vs invalid reasoning chains. Model parameters  are trained to minimize the binary cross entropy loss.

%% file: texfiles/experiments.tex
  \begin{table*}[]
    \centering
    \small{
    \begin{tabular}{@{}lcccccc@{}}
    {\bf Model} & \multicolumn{1}{c}{\bf Delexicalized} & \multicolumn{2}{c}{\bf Classification} & \multicolumn{2}{c}{\bf Ranking}\\
         & {\bf Representation} & {\bf F1}  & {\bf AUC-ROC} & {\bf P@1} & {\bf NDCG} \\

	& & (dev) test & (dev) test & (dev) test & (dev) test  \\
         \toprule
        \textsc{Retrieval} & n/a & (0.52) 0.50 & (0.75) 0.74 & (0.47) 0.47 & (0.59) 0.60  \\ 
        \textsc{Bert-QA} & n/a & (0.44) 0.43 & (0.52) 0.51 & (0.47) 0.47 & (0.48) 0.49 \\
        \textsc{Bert-chain} & No & (0.68) {\bf 0.64} & (0.88) {\bf 0.87} & (0.57) {\bf 0.55} & (0.65) {\bf 0.65} \\
        \textsc{Bert-grc} & Yes  & (0.63) {\bf 0.62} & (0.85) {\bf 0.85} & (0.55) {\bf 0.54} & (0.64) {\bf 0.64} \\        
        \multicolumn{2}{l}{Performance upper-bound:} & (1.00) 1.00 & (1.00) 1.00 & (0.76) 0.76 & (0.76) 0.76 \\ \bottomrule
    \end{tabular}
    }
      \caption{The ability of models to identify valid explanations (classification) or rank the set of explanations for each answer (ranking), with best test results highlighted. \textsc{Bert-grc} and \textsc{Bert-chain} perform better than \textsc{Retrieval} and \textsc{Bert-QA} methods, though fall short of the upper bound. Using the generalized chains (\textsc{Bert-grc}) performs similarly to \textsc{Bert-chain}, even though it is using less information (masking out overlapping noun phrases). }
    \label{tab:results}
\end{table*}

\section{Experiments}

For training, we use the annotated chains in the train split of \eqasc{} alongwith the `gold' chains provided in the QASC dataset (QSC gold chains are always considered valid reasoning chains).
We try two different ways of presenting chains to the model, namely
the original and generalized chain representations (GRCs), thus produce two models that we refer to as {\bf \textsc{Bert-chain}} and {\bf \textsc{Bert-grc}} respectively.
In earlier experiments, we did not find using chains for negative answer options (which are all invalid chains) to be useful 
(see Section~\ref{sec:negative-answer-option-chains}), so we use chains for correct answer options only. We use AllenNLP \cite{allennlp} toolkit to code our models.

We test on all the three proposed datasets. Since we are interested in finding explanations for the correct answer, we ignore the incorrect answer chains for the purpose of testing (they still accompany the dataset and can be used as additional training data since they are invalid reasoning chains by definition:  Section~\ref{sec:negative-answer-option-chains}). For \eqasc{} and \eobqa{}, we evaluate in two ways: First, treating the task as classification, we measure F1 and AUC-ROC (below). Second, treating the task as ranking the {\it set} of explanations for each answer, we measure P@1 and Normalized Discounted Cumulative Gain (NDMC) (also below). We use the trained model's probability of a chain being valid to rank the retrieved candidate chains for a given question and answer.

\subsection{Metrics}

{\bf \textsc{F1}} and {\bf \textsc{AUC-ROC}}: Viewing the task as classifying individual explanations, we report the area under the ROC (Receiver Operating Characteristics) curve, treating the valid explanation chains as the positive class. ROC curves are plots of true positive rate on the Y-axis against false positive rate on the X-axis. A larger area under the curve is better, with 1.0 being the best. 
Additionally, we report F1 for the positive class.

\noindent {\bf P@1} and {\bf \textsc{NDCG}}: 
Viewing the task as ranking the {\it set} of explanations for each answer,
P@1 measures the fraction of cases where the topmost ranked chain is a valid chain. This reflects the model's ability to find a
valid explanation for an answer, given the retrieval module.
Note that the upper bound for this measure is
less than 1.0 for \eqasc{}, as there are questions for which {\it none} of the
candidate chains are valid (discussed shortly in Section~\ref{retrieval-failure}).
NDCG (Normalized Discounted Cumulative Gain) measures how well ranked the candidates are when ordered by score, and is a measure widely used in the learning-to-rank literature.
Consider an ordered (as per decreasing score) list of N(=10) chains and corresponding labels $y_i \in \{0,1\}; i \in 1,2,..,N$, where $y_i=1$ represents a valid chain.
\textsc{NDCG} is defined per question (then averaged) as:
\begin{equation*}
    \text{NDCG} =  \frac{1}{Z}  \sum_{i=1}^{N} \frac{y_i} { \log_2(i+1) } 
\end{equation*}
where Z is a normalization factor so that perfect ranking score (when all the valid chains are ranked above all the invalid chains) is $1$. We define \textsc{NDCG} as $0$ if there are no valid chains.

 \begin{table*}[]
     \centering
     \small 
     \setlength\tabcolsep{2pt}    	
     \begin{tabular}{@{}llcc@{}}
\multicolumn{1}{c}{\bf Original chain} & \multicolumn{1}{c}{\bf Edited chain} & {\bf\textsc{Bert}} & {\bf\textsc{Bert}} \\
 &  & {\bf\textsc{-chain}} & {\bf\textsc{-grc}} \\
\toprule
{tadpole changes into a frog} & {tadpole changes into a frog} & & \\
{\bf AND} {the frog is a totem of \textit{metamorphosis}} & {\bf AND} {the frog is a totem of \textit{transformation}} & $0.21$ & $0.00$ \\
{\bf $\rightarrow$} { tadpoles undergo \textit{metamorphosis}} &  {\bf $\rightarrow$} { tadpoles undergo \textit{transformation}} & &  \\
\midrule
{insects can spread disease and destroy crops} & {insects can spread disease and decimate crops} & & \\
{\bf AND} {food crops are produced for local consumption} & {\bf AND} {food crops are produced for local consumption} & $0.11$ & $0.00$ \\
{\bf $\rightarrow$} {insects can destroy food} &  {\bf $\rightarrow$} { insects can decimate food} & &  \\
\bottomrule
     \end{tabular}
     \caption{Prediction Consistency: Examples from \eqascpert{} with changes in probability score (of being a valid reasoning chain) for different methods.
     Here, \textsc{Bert-grc} has (desirably) not changed its score due to an immaterial perturbation, while
     \textsc{Bert-chain} has, indicating greater stability for the GRC representation. This trend holds generally (Table~\ref{tab:stress_test}).
     \label{tab:stress_test_example}}
 \end{table*}


 \begin{table}[]

     \centering
     \small 
     \begin{tabular}{@{}ccccc@{}}
     \toprule
    \multicolumn{1}{c}{}  & {\bf \% cases with}   \\
    \multicolumn{1}{c}{\bf Model} & {\bf 0.0 change}   \\
    \midrule
        \textsc{Bert-chain} & $0.23\%$ \\ 
        \textsc{Bert-grc} & ${\bf 40.80\%}$ \\
        \bottomrule
     \end{tabular}
     
     \caption{Given an immaterial perturbation to a reasoning chain, a model's predicted probability of validity should not change if it is making consistent predictions. 
     We evaluate the absolute difference in probability scores of original and edited reasoning chains in \eqascpert{}. We observe that for $40.8\%$ and $0.23\%$ of the cases did not show any change in score for \textsc{Bert-grc} and  \textsc{Bert-grc} respectively. The results suggest that GRCs improve prediction consistency. 
     \label{tab:stress_test}}
 \end{table}

\subsection{Baselines}

We compare our model with two baselines, {\bf \textsc{retrieval}} and {\bf \textsc{Bert-QA}}.
Recall that our method first collects the top M candidate chains, ordered by retrieval score (Section~\ref{retrieval}).
Thus a simple baseline is to use that retrieval score itself as a measure of chain validity.
This is the {\bf \textsc{retrieval}} baseline.

We also consider a baseline, {\bf \textsc{Bert-QA}}, by adapting the approach of \newcite{perez2019finding} to our task.
In the original work, given a passage of text and a multiple choice question,
the system identifies the sentence(s) $S$ that are the most convincing evidence
for a given answer option $a_i$. To do this, it iteratively finds the sentence
that most increases the probability of $a_i$ when added to an (initially empty)
pool of evidence, using a QA system originally trained on the entire passage.
In other words, the probability increase is used as a measure of how
convincing the evidence sentence is. 
We adapt this by instead finding the {\it chain} that most increases
the probability of $a_i$ (compared with an empty pool of evidence),
using a QA system originally trained with all the 
candidate chains for $a_i$. For the QA system, we use the straightforward
BERT-based model described in \cite{khot2019qasc}. 
We then use that increase in probability of the correct answer
option, measured for each chain, as a measure of chain validity.
We call this baseline {\bf \textsc{Bert-QA}}.

\subsection{Results: Performance on eQASC \label{results:eQASC}}
The test results on the \eqasc{} are shown in Table~\ref{tab:results}. 
There are several important findings: \\
1. The best performing versions of \textsc{BERT-chain} and \textsc{BERT-grc} significantly outperforms the baselines.
In particular, the AUC-ROC is 11\% higher (absolute), NDCG rises from 0.60 to 0.64, and P@1 rises from 0.47 to 0.54 for \textsc{BERT-grc}, indicating substantial improvement.\\
2. The generalized chain representation does not lead to a significant reduction (nor gain) in performance,
despite abstracting away some of the lexical details through variabilization. This suggests the abstracted
representation is as good as the original, and may have some additional benefits (Section~\ref{consistency-results}). \\
3. The {\textsc{Bert-QA}} baseline scores surprisingly low.
A possible explanation is that, in the original setting, 
\newcite{perez2019finding}'s model learned to spot a
(usually) single relevant sentence among a passage of irrelevant
sentences. In our setting, though, all the chains are
partially relevant, making it harder for the model to
distinguish just one as central.

\begin{table}[t]
    \centering
    \small{
    \begin{tabular}{cccc}
     Model & {\bf P@1} & {\bf AUC-ROC} \\ \toprule
    \textsc{Retrieval} & $0.70$ & $0.58$  \\ %
    \textsc{Bert-chain} &  $0.85$  & $0.89$ \\  
    \textsc{Bert-grc} & $\bf 0.89$ & $0.86$ \\ 
         \bottomrule
    \end{tabular}
    }
    \caption{ Application of our (\eqasc{}-trained) model to a new dataset \eobqa{}. The high AUC-ROC figure suggests the model remains good at distinguishing valid from invalid chains. We report P@1 only for the questions which have at least one valid chain, i.e., where
    ranking is meaningful.}
    \label{tab:obqa}
\end{table}

\subsection{Results:Consistency in eQASC-perturbed \label{consistency-results}}

We posit that the generalized (GRC) chain representation may improve robustness to small changes in the chains, as the GRC abstracts away some of the lexical details. To evaluate this, we use the crowdworker-perturbed, (still) valid chains in \eqascpert{}. As the perturbed chain often follows the same/similar reasoning as the original one, this test can be considered one of consistency: the model's prediction should stay same.
To measure this, we record the model's predicted probability of a chain being valid, then compare these probabilities for each pair of original and perturbed chains. Ideally, if the model is consistent and the perturbations are immaterial, then these probabilities should not change.

The results are shown in Table~\ref{tab:stress_test}.
In a large fraction of the instances,  generalized chain representation exhibits no change. This is perhaps expected given the design of the GRC representations. Thus, using GRC not only achieves similar performance (Table~\ref{tab:results}), but produces more consistent predictions for certain types of perturbations. Table \ref{tab:stress_test_example} shows some examples.

\subsection{Results: Generalization to eOBQA}

We are also interested in the generality of the model, i.e., how well
it can transfer to a new dataset with no explanation training data
(i.e., the situation with most datasets). To measure this, we ran
our (\eqasc{}-trained) models on \eobqa{}, namely the annotated top-2 candidate chains
for OBQA test questions, to see if the models can still detect valid
chains in this new data.

The results are shown in Table~\ref{tab:obqa}, and again illustrate
that the BERT trained models continue to significantly outperform the retrieval baseline. High $P@1$ scores suggest that model is able to score a valid reasoning as the highest among the candidate whenever there is at least one such valid chain. The high AUC-ROC suggests that the model is able to effectively distinguish valid from invalid chains.

%% file: texfiles/analysis.tex
\section{Analysis and Discussions}

\subsection{GRC as Explicit Reasoning Rationale}
A potentially useful by-product of GRCs is that the underling reasoning
{\it patterns} are made explicit. For example, Figure~\ref{GRC-examples}
show some of the top-scoring GRCs. This may be useful
for helping a user understand the rationale behind a chain,
and a repository of high-scoring patterns may be useful as a
knowledge resource in its own right.
This direction is loosely related to certain prior works on inducing general semantic reasoning rules (such as \citet{tsuchida2011toward} who propose a method that induces rules for semantic relations based on a set of seed relation instances.) 

\begin{figure}
\centerline{
 \fbox{%
  \parbox{1\columnwidth}{\small
X can cause Y AND Y can start Z $\rightarrow$ X can cause Z \\
X is used for Y AND Z are X $\rightarrow$ Z are used for Y \\
X are formed by Y AND Y are made of Z \\
\hspace*{20mm} $\rightarrow$ X are formed by Z \\
X are Y AND Y are Z $\rightarrow$ X are Z \\
X produce Y AND Y is a Z $\rightarrow$ X produce Z \\
X increases Y AND X occurs as Z $\rightarrow$ Z increases Y \\
X changes Y AND Y is Z $\rightarrow$ X changes Z \\
X is Y AND X carries Z $\rightarrow$ Y carries Z \\
X changes an Y AND Z are examples of X $\rightarrow$ Z change an Y \\
X are formed by Y AND X are formed through Z \\
\hspace*{20mm} $\rightarrow$ Y can cause Z \\
X changes a Y AND Z start most X $\rightarrow$ Z can change Y 
}}}
\caption{Examples of some of the highest scoring generalized reasoning chains (GRCs) found in eQASC. \label{GRC-examples}}
\end{figure}

\subsection{Error Analysis} 
However, the \textsc{Bert-grc} model was not always able to correctly
distinguish valid from invalid GRC explanations. To better
understand why, we analyzed 100 scoring failures on \eqasc{} (dev),
looking at the top 50 chains (i.e., ranked as most valid by our model)
that were in fact annotated as invalid (false positives, FP), and
the bottom 50 chains (ranked most invalid) that were
in fact marked valid (false negatives, FN). We observed four main
sources of error: \\

\noindent
{\bf 1. Over-generalization:} ($\approx$ 45\% of the FP cases, $\approx$ 40\% of FN cases).
  Some generalized reasoning chains are merely plausible rather than a deductive proof,
  meaning that their instantiations may be annotated as valid or invalid depending
  on the context. For example,
  for the GRC
  \begin{myquote}
{\it X contains Y} AND {\it Z are in X} $\rightarrow$ {\it Z are in Y} 
\end{myquote}
  its instantation may have been marked as valid in 
  \begin{myquote}
{\it Cells contain nuclei} AND {\it Proteins are in cells} \\ 
\hspace*{10mm} $\rightarrow$ {\it Proteins are in nuclei}
 \end{myquote}
 but not for
 \begin{myquote}
  {\it Smog contains ozone} AND {\it Particulates are in} \\
  \hspace*{10mm} {\it smog} $\rightarrow$
  {\it Particulates are in ozone} 
  \end{myquote}
  (Ozone itself does not contain particulates).
  Here the context is important to the perception of validity, but has been lost in the generalized form. 

\noindent
    {\bf 2. Incorrect Declarative Form}:  (FP $\approx 20\%$, FN $\approx 30\%$).
    Sometimes the conversion from question + answer to a declarative 
    form $H$
    goes wrong, eg 
    \begin{myquote}
    {\it What do octopuses use ink to hide from? sharks}
    \end{myquote}
    was converted to the nonsensical sentence 
    \begin{myquote}
    {\it Octopuses do use sharks ink to hide from}.
    \end{myquote}
    In these cases, the annotations on chains supporting the original answer
    do not meaningfully transfer to the declarative formulation. (Here, FP/FN
    are due to label rather than prediction errors). 

\noindent
    {\bf 3. Shared Entity Detection}: (FP $\approx 10\%$, FN $\approx 10\%$)
    To detect and variabilize shared entities during GRC construction, we search for repeated noun phrases in the sentences.
    This operational definition of ``shared entities" can sometimes make mistakes, for example
    sometimes shared entities may be missed, e.g., {\it frog} and {\it bullfrog},
    or incorrectly equated due to stemming or wrong part of speech tagging, e.g., {\it organic} and {\it organism}.
    The resulting GRC may be thus wrong or not fully generalized, causing some errors. 

  \noindent {\bf 4. Model Failures}: (FP $\approx 25\%$, FN $\approx 10\%$)
    The remaining failures appear to be simply due the model itself,
    representing incorrect generalization from the training data.
    Additional training data may help alleviate such problems.

    Despite these, GRCs often abstract away irrelevant details, and may
    be worthy of further study in explanation research.

\subsection{Chains for Negative Answer Options \label{sec:negative-answer-option-chains}}

We also investigated whether we could skip using the eQASC annotations completely, and instead simply use the single QASC gold chains as positives, and chains for wrong answers as negatives (a form of distant supervision). However, we observed that but the results were significantly worse.
We also tried adding chains for wrong answers as additional negative examples to the full \eqasc{} dataset. However, we observed that this did not significantly improve (or hurt) scores. One possible reason for this is that \eqasc{} may already contain enough training signal. Another possible reason is that (invalid) chains for wrong answers may qualitatively differ in some way from invalid reasoning chains for right answers, thus this additional data does not provide reliable new signal.

\subsection{Limitations of Retrieval \label{retrieval-failure}}

Our focus in this paper has been on recognizing valid chains of reasoning,
assuming a retrieval step that retrieves a reasonable pool of candidates to start with (Section~\ref{retrieval}).
However, the retrieval step itself is not perfect: For QASC, designed so that at least one valid chain always exists, the retrieved pool of 10 contains no valid chains for 24\% of the questions (upper bound in Table~\ref{tab:results}), capping the overall system's performance.
To gauge the performance of our model when coupled with an improved retrieval system, we ran an experiment where, at test time, we explicitly add the gold chain to the candidate pool if it does not get retrieved (and even if there is some other valid chain already in the pool). We find the P@1 score rises from $0.54$ (Table~\ref{tab:results}) to $0.82$ (upper bound is now $1.0$). This indicates the model scoring algorithm is performing well, and that improving the retrieval system, e.g., by considering may more chains per question or modifying the search algorithm itself, is likely to have the biggest impact on improving the overall system.
Note also that the corpus itself is an important component: finding valid chains requires the corpus to contain a broad diversity of general facts to build chains from, hence expanding/filtering the corpus itself is another avenue for improvement.

\subsection{Future Directions}
The main purpose of this dataset is to generate explanations as an end-goal in itself, rather than improve QA scores (we do not make any claims in terms
of QA accuracy or ability to improve QA scores). Although much of NLP has focused on QA scores, more recent work has targeted explanation as an end-goal in itself, with ultimate benefits for tutoring, validation, and trust. Nonetheless, a useful future direction is exploring answer prediction and explanation prediction as joint goals, and perhaps they can benefit each other. 

Additionally, in the current work we have explored only a sequence of two sentences as an explanation for the third. Extending the proposed approaches for longer chains is an important future direction. 
We have proposed a technique for reducing reasoning chains to abstract chains. This technique makes assumptions about being able to match overlapping words. A future extension could explore more robust techniques for identifying abstract chains which do not make such assumptions.

%% file: texfiles/conclusions.tex
\section{Summary and Conclusion}

Explaining answers to multihop questions is important for understanding {\it why} an answer may be correct, but there is currently a dearth of suitable, annotated data. To address this, and promote progress in explanation, we contribute three new explanation datasets, including one with over 98k annotated reasoning chains - by far the largest repository of annotated, corpus-derived explanations to date. We also have shown this data can significantly improve explanation quality on both in-domain (QASC) and out-of-domain (OBQA) tasks. Finally, we have proposed and explored using a lightweight method to achieve a delexicalized representation of reasoning chains. While preserving explanation quality (despite removing details), this representation appears to be more robust to certain perturbations. 

%% file: appendix.tex
\section*{APPENDIX}

\section*{A. Additional Implementation Details}

\begin{itemize}
    \item Optimizer: We use Adam optimizer with initial learning rate of 2e-5
    \item Number of params: $\sim 110$M parameters (Bert-base uncased and classification layer)
    \item  Hyper-parameters: We search over following options for hyperparameter (1) one layer vs two layer classifier (2) negative class weight ( $(0.1,0.2,...,0.9)$) (3) using negative option chains or not; 
    for \textsc{Bert-grc} as well as \textsc{Bert-chain}. We perform model selection based on best dev split performance as per $P@1$. 
    \item Best model configuration for BERT-Chain: negative class weight = $0.2$; without using negative option chains; using a two layer classifier.
    Best model configuration for BERT-GRC: negative class weight = $0.3$; without using negative option chains; using a two layer classifier.
    \item We have uploaded code at \url{https://github.com/harsh19/Reasoning-Chains-MultihopQA}
\end{itemize}

\setlength{\belowdisplayskip}{4pt} \setlength{\belowdisplayshortskip}{4pt}
\setlength{\abovedisplayskip}{4pt} \setlength{\abovedisplayshortskip}{4pt} 


\begin{figure*}

  \noindent
      {\large {\bf B. Instructions to Crowdworkers}}

      \noindent
      Below are the instructions provided to the (Amazon Mechanical Turk) crowdworkers for chain annotation.
  
  \includegraphics[width=1.05\textwidth]{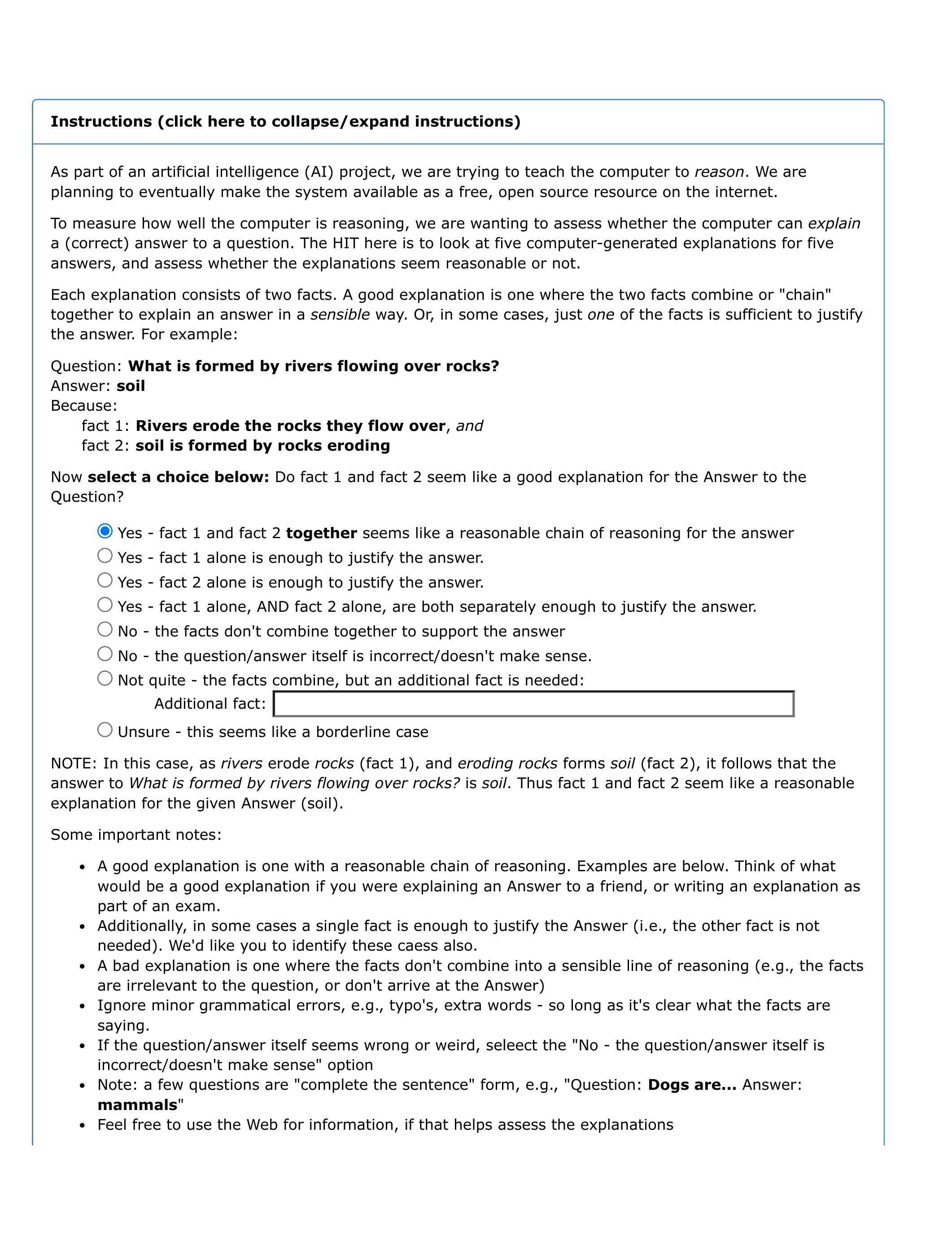}
\end{figure*}

\begin{figure*}
  \includegraphics[width=1.05\textwidth]{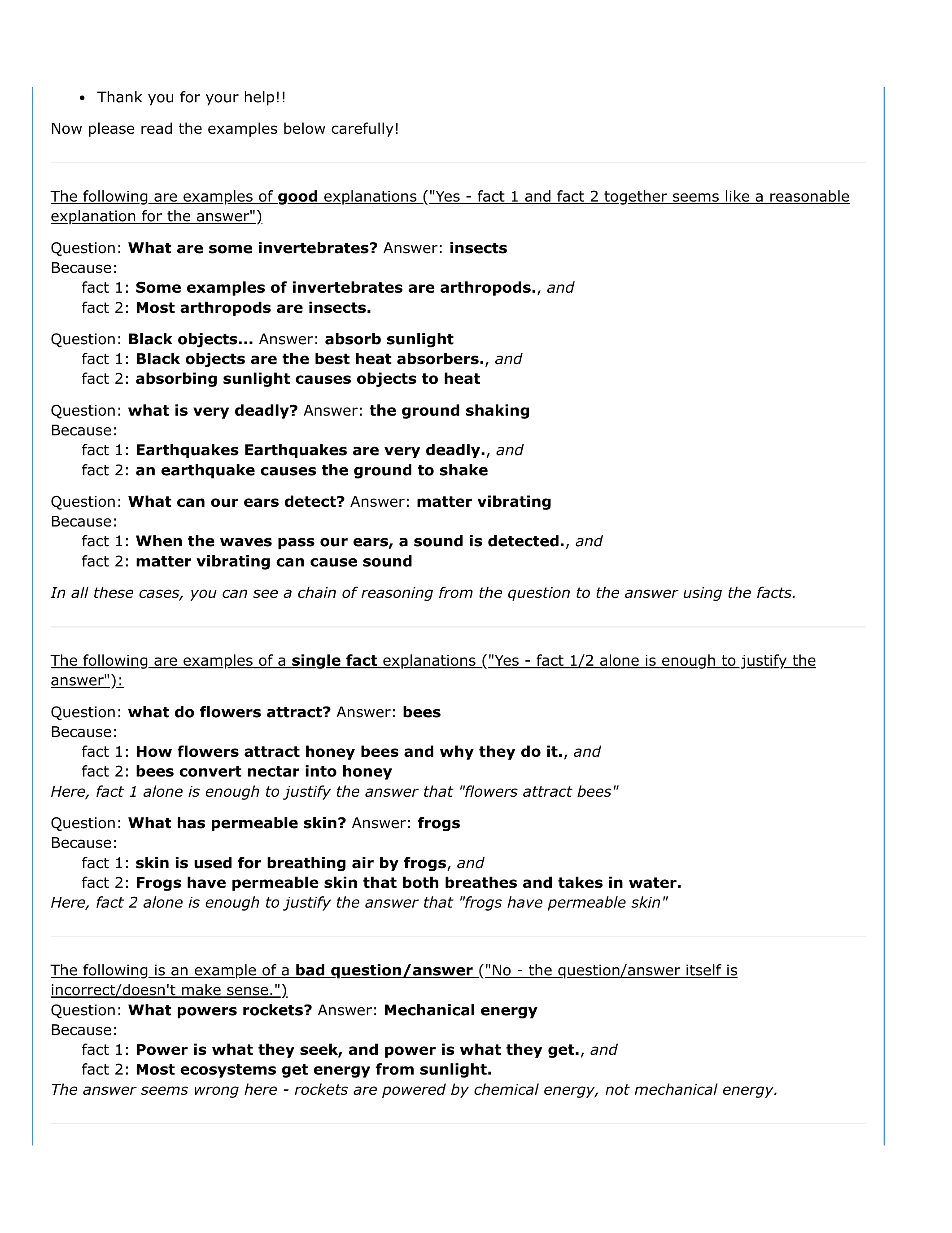}
\end{figure*}
\begin{figure*}
  \includegraphics[width=1.05\textwidth]{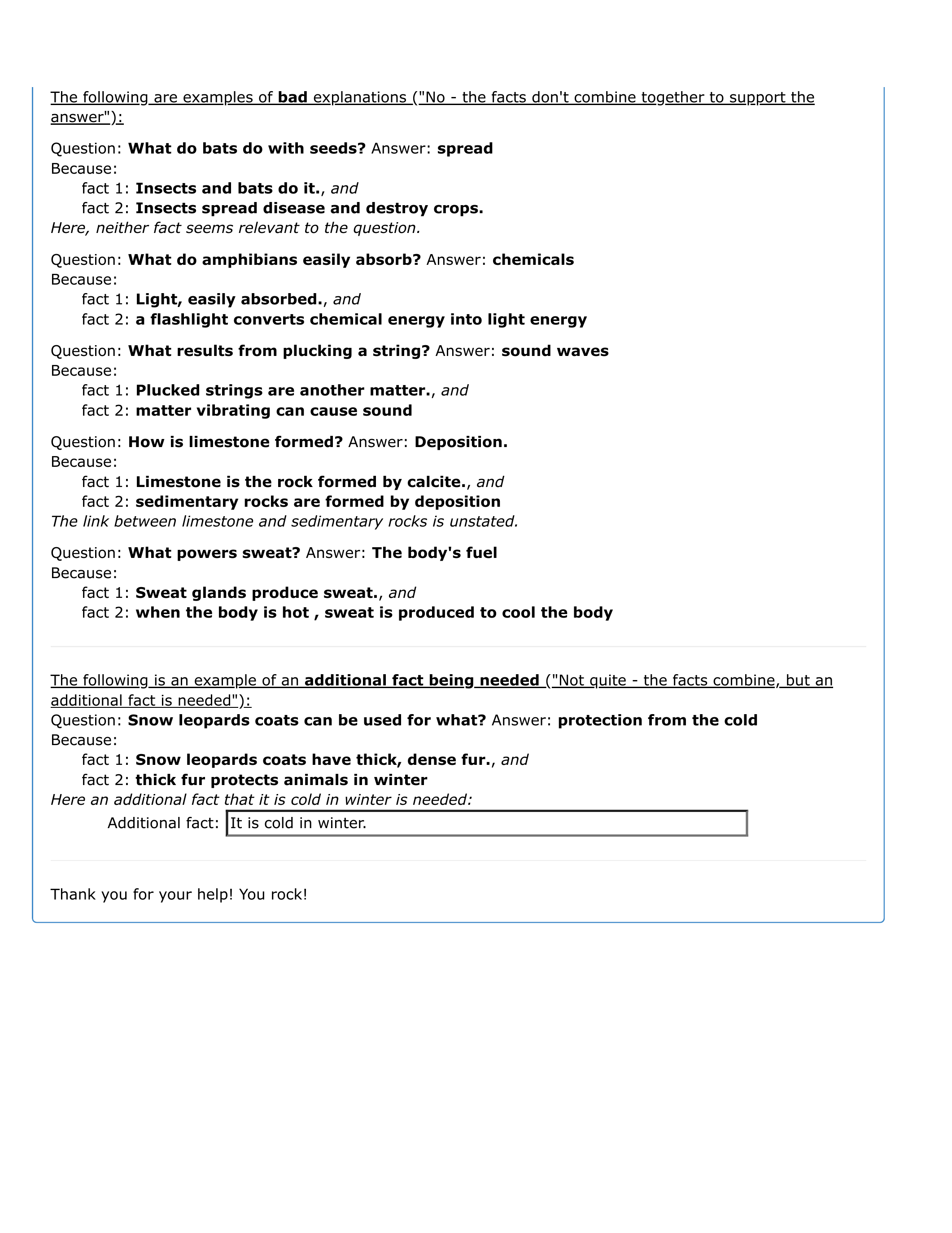}
\end{figure*}